%% file: main.tex
\documentclass[letterpaper, 10 pt, conference]{ieeeconf}  

\IEEEoverridecommandlockouts                              

\usepackage{graphics} 
\usepackage{epsfig} 
\usepackage{times} 
\usepackage{amsmath} 
\usepackage{amssymb}  
\usepackage{adjustbox}
\usepackage{multirow}
\usepackage{hhline}
\usepackage{array}
\usepackage{makecell}
\usepackage{bibentry}
\usepackage{epstopdf,cite}
\usepackage{xcolor}
\usepackage{color}
\usepackage{bbding}
\usepackage{pifont}

\usepackage{graphicx}
\usepackage{multicol}
\title{\LARGE \bf
PillarGen: Enhancing Radar Point Cloud Density and Quality via Pillar-based Point Generation Network
}

\author{Jisong Kim$^{1, \dagger}$, Geonho Bang$^{2, \dagger}$, Kwangjin Choi$^{2}$, Minjae Seong$^{2}$, \\ Jaechang Yoo$^{3}$, Eunjong Pyo$^{3}$, and Jun Won Choi$^{4, \ast}$
\thanks{$^{1}$Department of Electrical Engineering, Hanyang University, Seoul, 04763, Korea. {\footnotesize jskim@spa.hanyang.ac.kr}}
\thanks{$^{2}$Department of Artificial Intelligence, Hanyang University, Seoul, 04763, Korea. {\footnotesize \{ghbang,kjchoi,mjseong\}@spa.hanyang.ac.kr}}
\thanks{$^{3}$Radar Innovation Lab, HL Klemove, Seongnam, 13453, South Korea. {\footnotesize \{jc.yoo, eunjong.pyo\}@hlcompany.com}}
\thanks{$^{4}$Department of Electrical and Computer Engineering, College of Liberal Studies, Seoul National University, Seoul, 08826, Korea. {\footnotesize junwchoi@snu.ac.kr}}
\thanks{$\dagger$ Equally contributed authors}
\thanks{$\ast$ Corresponding author}
}

\begin{document}

\maketitle
\thispagestyle{empty}
\pagestyle{empty}


\begin{abstract}
In this paper, we present a novel point generation model, referred to as {\it Pillar-based Point Generation Network} (PillarGen), which facilitates the transformation of point clouds from one domain into another. PillarGen can produce synthetic point clouds with enhanced density and quality based on the provided input point clouds. The PillarGen model performs the following three steps: 1) {\it pillar encoding}, 2) {\it Occupied Pillar Prediction} (OPP), and 3) {\it Pillar to Point Generation} (PPG). The input point clouds are encoded using a pillar grid structure to generate pillar features. Then, OPP determines the active pillars used for point generation and predicts the center of points and the number of points to be generated for each active pillar. PPG generates the synthetic points for each active pillar based on the information provided by OPP. We evaluate the performance of PillarGen using our proprietary radar dataset, focusing on enhancing the density and quality of short-range radar data using the long-range radar data as supervision. Our experiments demonstrate that PillarGen outperforms traditional point upsampling methods in quantitative and qualitative measures. We also confirm that when PillarGen is incorporated into bird's eye view object detection, a significant improvement in detection accuracy is achieved.
\end{abstract}

\IEEEpeerreviewmaketitle
\section{INTRODUCTION}
In the field of robotics and autonomous driving, point cloud data acquired by ranging sensors such as LiDAR and radar are commonly used for 3D perception tasks. In general, the characteristics and distribution of point clouds vary based on sensor specifications and settings. For example, the sensor's quality and cost influence the density and intensity of point clouds, which in turn impacts perception performance. Hence, there may be situations where point clouds with specific characteristics need to be synthetically transformed into ones with different attributes. Specifically, using two sets of point cloud data, a model can be trained to generate synthetic point clouds that have the attributes of the target point clouds given the source point clouds available. 


\begin{figure}
\centering
\includegraphics[width=0.8\linewidth]{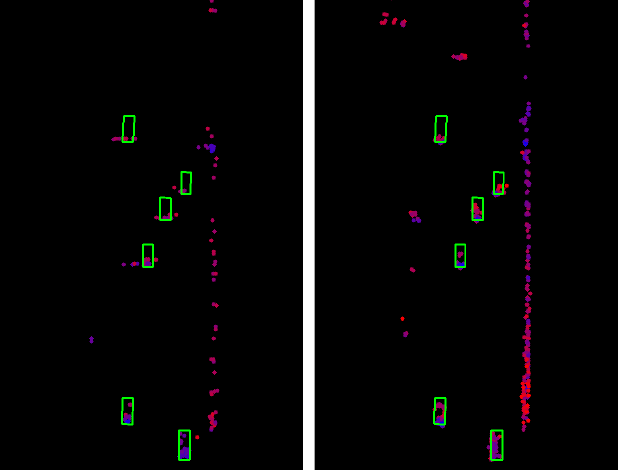}
\caption{\textbf{Radar Point Cloud Data Comparison.} Left: Sample from short-range radar. Right: Sample from long-range radar. The color of the points represents the radar cross-section (RCS) values, with lower values closer to blue and higher values closer to red, and the green boxes indicate the ground truth boxes.}
\label{fig:intro}
\end{figure}

This point generation task demonstrates its benefits when improving the quality of point clouds obtained from low-cost radar sensors, aiming to achieve performance levels comparable to those of high-resolution radar systems. Fig. \ref{fig:intro} shows two point clouds gathered from two different radar sensors, both representing the same scene. 
The right figure is long-range radar data which has a limited field of view but high density and intensity with minimal errors in the azimuth direction. The left figure is short-range radar data, which has poor azimuth resolution and is noisy. Notably, the short-range radar is more cost-effective than its long-range counterpart. Hence, it would be desirable to generate point clouds that mimic the attributes of a long-range radar. 
In this study, we aim to develop an effective point generation network that can transform point clouds from one domain to another. A primary challenge of this task is that the model should generate realistic point clouds that align closely with the true distribution of the target data while preserving the correspondence with the source point clouds. Similar challenges have been extensively explored in the field of generative image modeling \cite{Unpaired_I2I, Cond_I2I, stargan_I2I} and language translation \cite{S2S_LT, neural_LT, effective_LT}. However, the translation of point cloud data between domains still needs to be explored.

To date, various point generation models have been proposed to synthesize realistic point cloud data, such as \cite{zyrianov2022learning, UltraLiDAR, Nerf-lidar}. However, these approaches mainly focus on producing point cloud data within a single domain and lack the capability to generate target data based on source data conditionally. Another line of research relevant to our work is the point upsampling task, which aims to generate point clouds with increased density \cite{pu-net,pu-gan,pc2-pu,pugeo-net,dis-pu,mpu,pu-gcn}. While these models successfully produced high-density point clouds, they were mostly tailored for LiDAR point clouds capturing a single object in indoor settings. Our interest, however, is to deal with more complex scenes and noisy point clouds commonly obtained from real-world robotics and autonomous driving scenes. Point completion methods were proposed to augment incomplete 3D point cloud data to find a more complete representation of objects or scenes \cite{yang2018foldingnet, yuan2018pcn, nie2020skeleton, wen2020point, yu2021pointr}. However, these methods were tailored for refining object feature details in local regions rather than generating a complete set of point clouds.

In this paper, we propose a novel point generation model referred to as {\it Pillar-based Point Generation Network} (PillarGen). Our empirical study indicates that existing point upsampling methods struggle to accurately capture complex scenes within entire point clouds, failing to generate realistic point clouds in the target domain. This issue arises because many upsampling methods generate points by expanding feature vectors extracted from input points, leading to results closely mirroring the input domain. In contrast, PillarGen identifies specific regions for point generation and creates points within these areas, thereby enabling diverse and realistic point cloud generation across distinct domains.

PillarGen performs the following three steps: 1) {\it pillar encoding}, 2) {\it Occupied Pillar Prediction} (OPP), and 3) {\it Pillar to Point Generation} (PPG). Initially, raw radar points are transformed into pillar features through the {\it pillar encoding} \cite{pointpillars}. Then, 2D convolutional layers are employed to capture relationships between spatially adjacent pillars. Next, for each pillar grid, synthetic points are generated through OPP and PPG modules. OPP first identifies {\it active pillars} used to generate synthetic points selectively. Concurrently, it predicts both the average attributes and the number $K'$ of points to be generated for each pillar. Next, PPG produces $K'$ synthetic points for active pillars, adjusting their positions around a predicted center to determine final coordinates. Furthermore, for each synthetic point, PPG samples features from adjacent pillars using bilinear interpolation and predicts radar cross section (RCS) and velocity offsets additionally.

We evaluate our PillarGen on our proprietary radar datasets sourced from two distinct types of radars: two short-range radars and one long-range radar, all gathered on public roads. We trained PillarGen to enhance the density and quality of short-range radar data using the supervision of long-range radar data. Accounting for the unique attributes of radar, we introduce two evaluation metrics for our task: {\it Radar-specific Chamfer Distance} (RCD) and {\it Radar-specific Hausdorff Distance} (RHD). Our findings reveal that PillarGen can successfully synthesize radar point clouds whose distributions is close to that of the long-range radar data. The quality of data produced by PillarGen surpasses that of other point sampling methods. In addition, when the data generated by PillarGen is used for bird's eye view object detection, the detection accuracy is significantly improved compared to the case where PillarGen is not employed.

\section{RELATED WORK}

As various deep learning architectures for encoding point clouds, including PointNet \cite{pointnet}, PointNet++ \cite{pointnet++}, PointCNN \cite{pointcnn}, and DGCNN \cite{dgcnn}, emerged, several models for generating synthetic point clouds have also been proposed. 
Point generation networks were designed to learn the distribution of point cloud data and produce synthetic point clouds with distributions closely resembling real data \cite{fan2017point, achlioptas2018learning, valsesia2018learning, gadelha2018multiresolution, shu20193d, yang2019pointflow, han2019multi, luo2021diffusion, zyrianov2022learning, sun2020pointgrow}. These methods employed deep generative models such as generative adversarial networks (GANs) \cite{achlioptas2018learning, shu20193d, gadelha2018multiresolution, valsesia2018learning}, variational autoencoders (VAEs) \cite{fan2017point, han2019multi}, flow-based methods \cite{sun2020pointgrow, yang2019pointflow}, and diffusion methods \cite{luo2021diffusion, SDF_diffusion} to generate synthetic point clouds.


Point cloud completion methods generate and fill in missing parts of point clouds. They primarily concentrate on accurately reconstructing a subset of point clouds based on semantic information inferred from the input point clouds. Point cloud completion methods can be categorized into two approaches: voxel-based methods \cite{dai2017shape, nguyen2016field, wang2021voxel} and point-based methods \cite{wen2021cycle4completion, wen2020point, tang2022lake}. The voxel-based methods obtain a fixed-size 3D feature map by converting unordered points via voxelization, which is then processed using 3D CNNs or differential gridding layers to complete the point cloud \cite{han2017high, xie2020grnet}. Point-based methods leverage point encoding methods such as PointNet \cite{pointnet}, PointNet++ \cite{pointnet++}, or attention layers \cite{vaswani2017attention} to generate features, which are then utilized within an encoder-decoder framework to complete the point cloud \cite{yuan2018pcn,chen2023anchorformer,yu2021pointr, zhou2022seedformer}.

\begin{figure*}[t]
    \centering
        \includegraphics[width=0.9\linewidth]{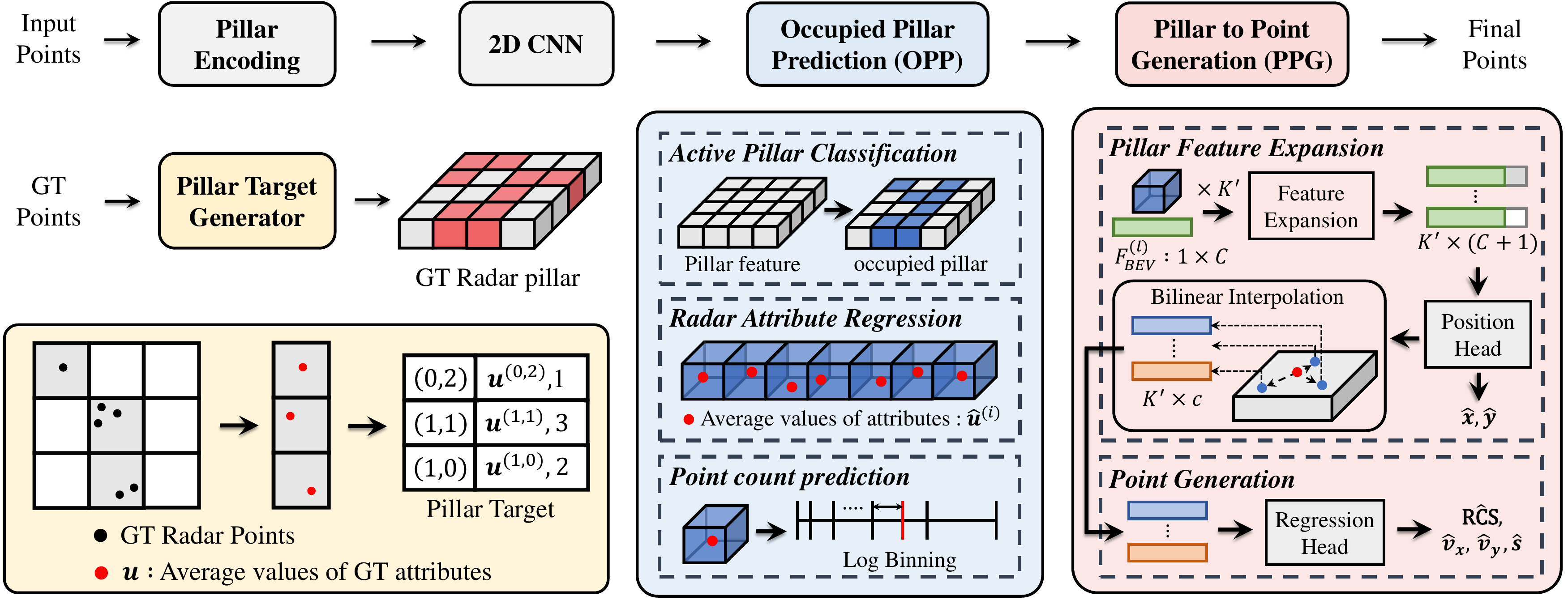 }
        \caption{\textbf{Overall architecture of the proposed PillarGen:} PillarGen takes short-range radar points as input to generate long-range radar points. The Pillar Target Generator forms GT Radar pillars using GT radar points. Pillar Encoding and 2D CNN modules extract BEV features from input radar points. OPP module uses BEV features and GT Radar pillars to simultaneously conduct three predictions. Finally, PPG module generates long-range radar points through pillar feature expansion and point generation.}
    \label{fig:overall}
\end{figure*}


Point cloud upsampling methods increase the resolution of input point clouds by interpolating 3D point data. These methods generate points based on inherent attributes of input point clouds, such as shape information and priors on smoothness. Some work \cite{pu-net, pu-gan, pu-gcn, pu-transformer, mpu} extract features using diverse point encoding methods and generate synthetic points by expansion modules like multi-step upsampling or node shuffle. In contrast, other approaches \cite{dis-pu, pugeo-net, pc2-pu} initially produce a coarse point cloud, which is then refined to reduce the positional error of the points.

Both point cloud completion and upsampling methods generate additional points that conform to the distribution of the input point clouds. However, our challenge with PillarGen can be more demanding than these tasks because PillarGen aims to transform point clouds from one domain to another, where the distribution in the source domain may not necessarily align with that in the target domain. In this study, our study particularly focuses on scenarios where the quality of point clouds in the source domain is inferior to that in the target domain.



\section{Proposed PillarGen}
In this section, we present the details of the proposed PillarGen. Fig. \ref{fig:overall} depicts the overall structure of PillarGen. 


\subsection{Pillar Encoding}
\label{subsec: encoding}

We first adapt the pillar feature extraction method introduced in \cite{pointpillars} for encoding our radar point clouds. Unlike LiDAR points, radar points are composed of 3D coordinates $(x, y, z)$, RCS, and Doppler velocity $(v_x, v_y)$. Therefore, we modify the existing pillar encoding module to utilize all radar point attributes $(x, y, z, RCS, v_x, v_y)$  as an input and generate the corresponding pillar features. These features are then passed through a pillar scatter module to obtain a pseudo-image. We utilize a 2D CNN-based encoder with convolutional blocks using 1x, 2x, and 4x stride to generate multi-scale {\it Bird's Eye View} (BEV) feature maps. Finally, these multi-scale features are concatenated to generate a BEV feature $F_{BEV}$ $\in \mathbb{R}^{C \times H \times W}$. As illustrated in Fig. \ref{fig:overall}, we also employ the {\it Pillar Target Generator} to generate a pseudo-image from a target radar data. 
Each element of the pseudo-image contains the arithmetic mean $({x_c}, {y_c}, {RCS_c}, {v_{xc}}, {v_{yc}})$ of point attributes. It also contains the total number of points $K$ within each pillar for the target radar data.
Utilizing the indices of these non-empty pillars, the six-dimensional vectors $({x_c}, {y_c}, {RCS_c}, {v_{xc}}, {v_{yc}}, K)$ are scattered back to their corresponding original positions in pillar grids, resulting in a pseudo-image with dimensions \( (6, H, W) \).

\subsection{Occupied Pillar Prediction (OPP)}
\label{subsec: OPP}

OPP module, as shown in the blue bounding box in Fig. \ref{fig:overall}, is designed to simultaneously predict three distinct values based on the $F_{BEV}$ by applying a $1\times1$ convolution layer. 
Initially, the {\it active pillar classification} module predicts probability $p_{occ}$ indicating whether a given pillar contains target radar points. Pillars containing these target radar points are termed as {\it active pillars}. Note that pillars with \( p_{\text{occ}} > 0.1 \) are identified as active.
Next, the {\it Radar Attribute Regression} module estimates the average attribute values of the points in each pillar. The predicted average attribute is denoted as $\hat{\mathbf{u}}=(\hat{x}_c, \hat{y}_c, \hat{RCS}_c, \hat{v}_{xc}, \hat{v}_{yc})$. Lastly, the {\it point count prediction} module predicts the number of target radar points contained within each pillar. This predicted value corresponds to the number $K'$ of points to be generated by the subsequent PPG module. Although a common approach to predict the number of points is performing regression for each pillar, we found that the prediction accuracy varies with the range of point numbers, owing to the long-tailed distribution of the point counts. Consequently, we address this issue by devising a log-scale binning approach. Log-scale binning is defined as 
\begin{equation}\label{bin_target}
    \begin{aligned}
         & \text{bin}^{(i)} =\lfloor{\log_2{K^{(i)}} }\rfloor, \\[1pt]
         & \text{res}^{(i)} = \log_2(K^{(i)} - 2^{\text{bin}^{(i)}} + 1), \\[1pt]
    \end{aligned}
\end{equation}
where \( \text{bin}^{(i)} \) denotes the sub-interval that discretely partitions the range for the number of points for the $i$-th pillar and \( \text{res}^{(i)} \) serves as the residual used for more accurate prediction within that sub-interval. In our approach, we utilize a classifier network to choose the most probable bin. This initial estimate is then refined by predicting the residual through a regression network, producing the final estimate $K'$.



\subsection{Pillar to Point Generation (PPG)}
\label{subsec: PPG}

PPG generates the synthetic points for the active pillars based on the information provided by OPP. First, to generate $K'_l$ synthetic points in the $l$-th active pillar, the {\it pillar feature expansion} module replicates the BEV feature vector ${F_{BEV}^{'(l)}}$ $K'_l$ times.
We append a random number to each of the $K'_l$ duplicated features to prevent the generation of identical points \cite{mpu}. 
Then, based on the duplicated features, the {\it position head} predicts the position offsets for $K'_l$ points to be generated. Two {\it Multi-Layer Perceptron (MLP)} layers are applied to the duplicated features to predict 2D coordinate offsets $(\Delta{x}, \Delta{y})$. These offsets are added to the position values $(\hat{x}_c, \hat{y}_c)$ predicted by OPP, producing the 2D coordinates $(\hat{x}, \hat{y})$. Next, using the predicted coordinates of each point, we apply bilinear interpolation on $F_{BEV}$ to obtain the interpolated pillar features $F^{(l)} \in \mathbb{R}^{K'_l \times C}$. 
The {\it regression head} applies two MLP layers to the interpolated features \( F^{(l)} \) to predict the offsets $(\Delta{RCS}, \Delta{v}_{x}, \Delta{v}_{y})$ for RCS and velocity. These offsets are also added to the average attributes $(\hat{RCS}_c, \hat{v}_{xc}, \hat{v}_{yc})$ to produce $(\hat{RCS}, \hat{v}_{x}, \hat{v}_{y})$.
Finally, we predict the confidence score for each point generated. Combined with the occupancy probability \( p_{occ} \), the confidence score indicates the uncertainty of each point generated. For every point, the regression head predicts the score offset $\Delta s$ and adds it to \( p_{occ} \) to produce the final confidence score, $\hat{s}$. 

In summary, PPG predicts the six attributes \( ( \hat{x} \), \( \hat{y} \), \( \hat{RCS} \), \( \hat{v_x} \), \( \hat{v_y} \), \( \hat{s} ) \) for each point generated from the $l$-th active pillar. Points generated from each active pillar are then merged to produce the complete set of predicted points. 


\subsection{Loss function}
The total loss function $\textit{L}_{total}$ is given by
\begin{align}\label{total_loss}
    \textit{L}_{total}= \textit{L}_{opp} + \textit{L}_{ppg},
\end{align}
where \(L_{\text{opp}}\) is OPP module's loss function and \(L_{\text{ppg}}\) is PPG module's loss function. 
First, \(L_{\text{opp}}\) is given by
\begin{align}\label{opp_loss}
    \textit{L}_{opp} = \textit{L}_{opp\text{-}cls} + \textit{L}_{opp\text{-}reg} + \textit{L}_{opp\text{-}bin},
\end{align}
where ${L}_{opp\text{-}cls}$ represents the classification loss using the focal loss \cite{focal} for the {\it active pillar classification} and ${L}_{opp\text{-}reg}$ is the smooth \(L1\) loss for the {\it Radar Attribute Regression}. $\textit{L}_{opp\text{-}bin}$ uses both the focal loss and smooth \(L1\) loss for training the {\it point count prediction}. 

The loss function $\textit{L}_{pgg}$ for PPG quantifies the distance between the generated points and the ground truth (GT) points, i.e., 
\begin{align}\label{ppg_loss}
    \textit{L}_{ppg} = \textit{L}_{ppg\text{-}local} + \textit{L}_{ppg\text{-}global},
\end{align}
where ${L}_{ppg\text{-}local}$ computes the distance loss between point sets within each pillar, and ${L}_{ppg\text{-}global}$ evaluates the distance loss across the entire point set. For ${L}_{ppg\text{-}local}$, we categorize each predicted active pillar set $\mathcal{\hat{V}}$ into two categories based on their alignment with GT active pillar set $\mathcal{V}$. If $\mathcal{\hat{V}}$ aligns with any pillar in $\mathcal{V}$, we label it as a positive pillar set $\mathcal{\hat{V}}_{\text{pos}}$ and its corresponding GT pillar as $\mathcal{V}_{\text{pos}}$. Otherwise, it is labeled as a negative pillar set $\mathcal{\hat{V}}_{\text{neg}}$. We compute the distance between points within each pair of positive pillars as
\begin{align}
D^{(m)}_{(j)} = \min_{\hat{\mathbf{p}}_{xy}\in \hat{\mathcal{V}}_{pos}^{(m)}}||\mathbf{p}_{xy}^{(j)} - \hat{\mathbf{p}}_{xy}||_2^2,
\end{align}
where \(D_{(j)}^{(m)}\) is the minimum L2 distance from the \(j\)-th point's 2D location represented by coordinates \(\mathbf{p}_{xy}^{(j)}\) in \(\mathcal{V}_{pos}^{(m)}\) to any point coordinates \(\hat{\mathbf{p}}_{xy}\) in \(\mathcal{\hat{V}}_{pos}^{(m)}\). Here, \( m = 1, 2, \dots, |\hat{\mathcal{V}}_{\text{pos}}| \)
serves as the index for each positive pillar, while \( j = 1, 2, \dots, |\mathcal{V}_{\text{pos}}^{(m)}| \) is the index for each point within the $m$-th non-empty GT pillar.
Additionally, we utilize the set of distances to generate target values for distance confidence score as
\begin{align}
s_{(j)}^{(m)} = \min\left(1, \frac{d_{\text{std}}}{\sqrt{D^{(m)}_{(j)}}}\right),
\end{align}
where $d_{std}$ is set to 0.25 as the acceptable error distance. If a generated point does not have a corresponding match in $\mathcal{V}_{pos}^{(m)}$ or is within $\hat{\mathcal{V}}_{\text{neg}}$, its confidence score is set to zero. The ${L}_{ppg\text{-}local}$ is formulated as
\begin{align}
{L}_{ppg\text{-}local} =\frac{1}{|\mathcal{V}_{pos}|}({L}_{\text{2D}} + {L}_{\text{feat}}) + \frac{1}{|\hat{\mathcal{V}}|}({L}_{\text{score}}),
\end{align}
where
\begin{align}
&{L}_{\text{2D}} = \sum_{m}\frac{1}{|\mathcal{V}_{pos}^{(m)}|}\sum_{j}D_{(j)}^{(m)},\\
&{L}_{\text{feat}} =\sum_{m}\frac{1}{|\mathcal{V}_{pos}^{(m)}|}\sum_{\substack{(j, k)\in I^{(m)} \\  \text{attr} \in \{RCS, v_x, v_y\}}}|p_{\text{attr}}^{(j)} -\hat{p}_{\text{attr}}^{(k)}|,\\
&{L}_{\text{score}} = \sum_{l}\frac{1}{|\hat{\mathcal{V}}^{(l)}|} \mathcal{F}_{\text{BCE}}(\mathbf{s}^{(l)}, \hat{\mathbf{s}}^{(l)}),
\end{align}
where \( l = 1, 2, \dots, |\hat{\mathcal{V}}| \) is an index for each predicted active pillar and $I^{(m)}$ consists of index pairs \( (j, k) \). Here, $k$ represents the index of the closest generated point to the $j$-th point in \( \hat{\mathcal{V}}_{pos}^{(m)} \). $\mathcal{F}_{\text{BCE}}$ denotes Binary Cross Entropy loss. For ${L}_{ppg\text{-}global}$, we utilize a modified Chamfer Distance, denoted as \( \text{RCD}_{5D} \), to account for both spatial and radar-specific features differences across the point clouds. \(L_{ppg\text{-}global}\) is obtained by computing \( \text{RCD}_{5D} \) between the entire set of predicted points and the corresponding GT points. The detailed definition of \( \text{RCD}_{5D} \) is provided in Eq.~\ref{eq:RCD_5D}.

\section{EXPERIMENTS}

\subsection{Experimental Setup}
\label{subsec: exp_setup}
\subsubsection{Datasets} We use a self-collected dataset to train and evaluate the efficacy of our PillarGen model. This dataset consists of 251 driving sequences, with 51 sequences gathered from highways and 200 sequences from urban roads. Each sequence comprises the synchronized data acquired from six multi-view cameras, two short-range radars, and a single long-range radar. Two short-range radar sensors are positioned at the vehicle's front corners, and a long-range radar sensor is located at the front. The radar data was sampled at a frequency of $16.2 Hz$, resulting in a total of 125,790 samples. The range of point clouds used for processing was set to $0$-$100$ meter. The field of view of the short-range radars was set to $\pm 40^\circ$ for the $0$-$60$ meter range. The field of view of the long-range radars was $\pm 12.5^\circ$ for the $60$-$100$ meter range. 


\subsubsection{Evaluation metrics} 
Chamfer Distance (CD) and Hausdorff Distance (HD) are commonly used metrics to measure the difference between two distinct point clouds. However, these metrics were mainly developed for point clouds represented as 3D spatial coordinates, and they do not cater to radar-specific attributes like \((RCS, v_x, v_y)\). Hence, we propose the {\it Radar-specific Chamfer Distance} (RCD) and {\it Radar-specific Hausdorff Distance} (RHD). We present four variations, \( \text{RCD}_{2D}, \text{RHD}_{2D}, \text{RCD}_{5D}, \) and \( \text{RHD}_{5D} \). 
First, \( \text{RCD}_{2D} \) and \( \text{RHD}_{2D} \) can be obtained by utilizing only the $(x, y)$ 2D coordinates for CD and HD, rather than the full 3D coordinates. This is because the accuracy of the $z$ coordinate is not critical for radar sensors. Before discussing the \( \text{RCD}_{5D} \) and \( \text{RHD}_{5D} \), we need to define the distance between two radar points \(a\) and \(b\) as follows
\begin{align}
d_{2D}(a, b) &= \| a_{2d} - b_{2d} \|_2^2 ,\\
d_{attr}(a, b) &= \| a_{attr} - b_{attr} \|_1 ,
\end{align}
where $a_{2d}$ and $b_{2d}$ indicate the $(x, y)$ components of points \(a\) and \(b\) respectively. Meanwhile, $a_{attr}$ and $b_{attr}$ denote the \( (\text{RCS}, v_x, v_y) \) components of points \(a\) and \(b\). Given two point sets \(P\) and \(Q\), the index \(j^*\) of the closest point in \(Q\) for each point \(p_i \in P\), and the index \(i^*\) of the closest point in \(P\) for each point \(q_i \in Q\), are obtained from
\begin{equation}\label{index_j}
j^* = \arg\min_{q_j \in Q} d_{2D}(p_i, q_j),
\end{equation}
\begin{equation}\label{index_i}
i^* = \arg\min_{p_i \in P} d_{2D}(q_j, p_i).
\end{equation}
With these definitions we can express the metrics \( \text{RCD}_{5D} \) and \( \text{RHD}_{5D} \) as
%
\begin{equation}
    \begin{aligned}\label{eq:RCD_5D}
        &\text{RCD}_{5D}(P, Q) = \frac{1}{|P|} \sum_{p_i \in P} \left( d_{2D}(p_i, q_{j^*}) + d_{attr}(p_i, q_{j^*}) \right) \\
        & \;\;\; + \frac{1}{|Q|} \sum_{q_j \in Q} \left( d_{2D}(q_j, p_{i^*}) + d_{attr}(q_j, p_{i^*}) \right),
    \end{aligned}
\end{equation}
\begin{equation}
    \begin{aligned}
        &\text{RHD}_{5D}(P, Q) = \max \left\{ \max_{p_i \in P} \left( d_{2D}(p_i, q_{j^*}) + d_{attr}(p_i, q_{j^*}) \right) , \right. \\
        & \;\;\; \qquad \left. \max_{q_j \in Q} \left( d_{2D}(q_j, p_{i^*}) + d_{attr}(q_j, p_{i^*}) \right) \right\}
    \end{aligned}
\label{eq:RHD_5D}
\end{equation}
where \( |P| \) and \( |Q| \) denote the number of points in \( P \) and \( Q \), respectively.

\input{table/SOTA_val}
\input{table/ablation_main}
\input{table/ablation_global_local}
\input{table/ablation_detection}

\begin{figure*}[t]
  \hspace{1.5cm} Input \hspace{1.3cm} PU-NET \cite{pu-net} \hspace{0.6cm} 
  PU-GCN \cite{pu-gcn} \hspace{0.7cm} 
  Dis-PU \cite{dis-pu} \hspace{1.3cm} 
  Ours \hspace{2.0cm} GT \\[3pt]
  \includegraphics[width=0.99\linewidth]{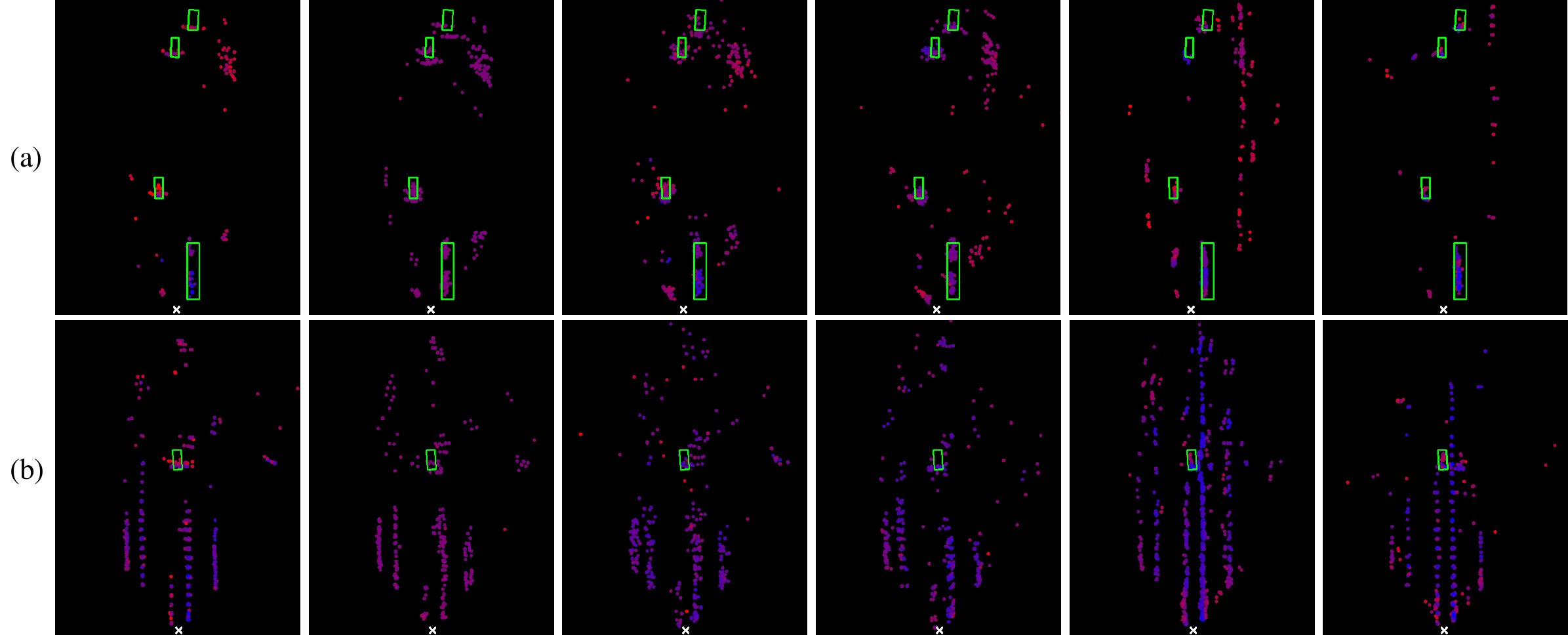}
  \caption{\textbf{Qualitative Results on our test dataset.} We compare the point clouds generated from short-range radar points using different method (PU-NET, PU-GCN, Dis-PU, and ours) against long-range radar points. The color of the points represents the RCS values, with lower values closer to blue and higher values closer to red and the green boxes indicate the ground truth boxes.}
  \label{fig:qualitative_results}
\end{figure*}

\subsubsection{Implementation Details} We used a multi-sweep method, similar to \cite{GrifNet}, to increase the point density. As described in Section \ref{subsec: encoding}, we implemented our model on a modified PointPillars \cite{pointpillars} architecture and set pillar size to $(2.5m, 2.5m)$. We first trained OPP module and backbone for 100 epochs with batch size 128 to obtain pre-trained weights. With these weights, we initialized the end-to-end model and conducted training for additional 100 epochs with a batch size of 128. In both training phases, we used the Adam \cite{adam} optimizer with a OneCycle learning rate schedule. The learning rate was set at 0.001, with a weight decay of 0.01 and momentum of 0.9. In inference, PillarGen generates points with a confidence score $\hat{s}$ is greater than $0.1$ to reduce points generated in inaccurate locations.

\subsubsection{Baselines} In order to compare our PillarGen, we utilized three baseline models: PU-NET \cite{pu-net}, PU-GCN \cite{pu-gcn}, and Dis-PU \cite{dis-pu}. 
We used 420 points obtained from short-range radar as an input and treated 840 points obtained from long-range radar as a target radar data. 
We trained baseline models for 100 epochs with batch size 64. Apart from these, we followed the default settings of each model.

\subsection{Quantitative Results}
Table~\ref{table:sota_val} presents a performance comparison of our PillarGen with other methods on our test dataset. PillarGen outperforms all other competitive methods in all evaluation metrics. These results show that pillar-based point generation method can effectively consume complex scene contexts from input radar points and generate points that follow the target distribution but retain the scene context. Moreover, our model not only generates spatially accurate points but also accurately predicts the radar-specific features for each point.

\subsection{Qualitative Results}
Figure \ref{fig:qualitative_results} presents a comparison of the qualitative results obtained from PillarGen with those of other point upsampling methods. Our model demonstrates the ability to generate densely populated points that closely match the distribution of ground truth points, while notably reducing the presence of noise points compared to other methods. Additionally, by examining the RCS values depicted by point colors, it becomes apparent that our model outperforms comparative models in predicting radar-specific features.

\subsection{Ablation Studies}
\subsubsection{Component Analysis}
Table \ref{table:ablation_main} provides an analysis of each component within our model. In our baseline, we solely employed the Active Pillar Classification within the OPP module, resulting in the generation of a fixed number of $8$ points per pillar. During the point prediction stage within PPG, it utilized the centroid of the pillar for generating points and did not predict a confidence score. We can observe that adding Log Binning to the baseline model, which adaptively determines the number of points generated, improves the performance across all metrics. Radar Attribute Regression, which predicts the centroid of points within a pillar and regresses the offset against the predicted centroid, further improves the performance of RCD (by 0.64 at $\text{RCD}_{2D}$ and 3.16 at $\text{RCD}_{5D}$). 
Finally, adding the confidence score leads to a gain across all metrics.

\subsubsection{Effects of Global and Local Loss}
We conducted experiments to compare the performance of two loss functions. We propose the global loss that compares points generated within positively aligned pillars. The objective is to refine the point distribution in these selected pillars to closely match the global structural characteristics of the GT point cloud. Compared to this, the local loss, which compares points generated at the pillar-level with their corresponding GT points, can generate fine-grained points in small regions. As shown in Table ~\ref{table:ablation_global_local}, the local loss which can generate refined points, yields higher performance. Also, by leveraging the complementary characteristics of both losses, we achieved the best performance in terms of the RCD and RHD.

\subsection{Application to BEV Object Detection}
We further apply our method to BEV Object Detection to demonstrate the extensibility of our model. Performance evaluations were conducted across three classes including vehicle, large-vehicle, and pedestrian. Specifically, the categories vehicle and large-vehicle were differentiated based on a $5m$ length criterion. The evaluation metrics from the nuScenes dataset \cite{nuscenes} were adopted for this application. As shown in Table ~\ref{table:ablation_detection}, our model significantly outperforms models relying solely on short-range radar points across all classes, with an impressive increase in mean Average Precision (mAP) by $3.16$. This demonstrates our model's effectiveness in point cloud generation, as well as its ability to enhance BEV object detection capabilities. 

\section{CONCLUSIONS}
In this paper, we proposed PillarGen, a novel pillar-based point generation network for translating point clouds from one domain to another. The proposed PillarGen uses a pillar structure to encode input point clouds and generate synthetic ones. First, OPP determines the active pillars where points should be generated and predicts attributes such as the center of points and the number of points to be generated. Based on this information, PPG generates the synthetic points for each active pillar. PPG also utilizes a confidence score to remove inaccurately positioned points, ensuring that the synthetic points closely resemble the target points. The quantitative and qualitative results demonstrated that PillarGen outperformed conventional point upsampling models by significant margins. Moreover, when incorporated into BEV object detection, PillarGen improved detection accuracy.

\section{ACKNOWLEDGMENT}
This work was supported by Institute for Information \& communications Technology Promotion(IITP) and the National Research Foundation of Korea (NRF) grant funded by the Korea government (MSIT) (No.2021-0-01314,Development of driving environment data stitching technology to provide data on shaded areas for autonomous vehicles and 2020R1A2C2012146)

\bibliographystyle{plain}
\bibliography{Reference}

\end{document}

%% file: table/SOTA_val.tex
\newcolumntype{C}{>{\centering\arraybackslash}p{2.3em}}
\newcolumntype{'}{!{\vrule width 0.1pt}}
\renewcommand{\arraystretch}{1.0}

\begin{table}[t]
\caption{Quantitative comparisons with other methods.}
\label{table:sota_val}
\begin{center}
\begin{adjustbox}{width=0.45\textwidth}
{
\fontsize{7pt}{9pt}\selectfont
\begin{tabular}{c 'c c 'c c}
\Xhline{3\arrayrulewidth} 
Method & \( \text{RCD}_{2D}\downarrow \) & \( \text{RHD}_{2D}\downarrow \) & \( \text{RCD}_{5D}\downarrow\) & \( \text{RHD}_{5D}\downarrow \) \\[2pt]
\hline 
\Xhline{0.1\arrayrulewidth}

PU-Net \cite{pu-net} & 23.67 & 549.82 & 29.32 & 555.47 \\
PU-GCN \cite{pu-gcn} & 18.38 & 485.65 & 21.83 & 489.10 \\ 
Dis-PU \cite{dis-pu} & 17.70 & 496.85 & 21.08 & 500.23 \\

\Xhline{0.1\arrayrulewidth}

\textbf{Ours} & \textbf{13.92} & \textbf{417.49} & \textbf{16.67} & \textbf{420.24} \\
\Xhline{3\arrayrulewidth}
\end{tabular}
}
\end{adjustbox}
\end{center}
\end{table}

%% file: table/ablation_main.tex
\renewcommand{\arraystretch}{1.0}

\begin{table}[t]
\caption{Ablation study of our components. LB: Log Binning method. RAR: Radar Attribute Regression method. $s$: confidence score}
\label{table:ablation_main}
\begin{center}
\begin{adjustbox}{width=0.45\textwidth}
{
\fontsize{10pt}{12pt}\selectfont
\begin{tabular}{ccc|cccc}
\Xhline{4\arrayrulewidth}
\multirow{2}{*}{LB} & \multirow{2}{*}{RAR} & \multirow{2}{*}{$s$} & \multicolumn{4}{c}{Performance} \\ \cline{4-7}
&  &  & $\text{RCD}_{2D}\downarrow$ & $\text{RHD}_{2D}\downarrow$ & $\text{RCD}_{5D}\downarrow$ & $\text{RHD}_{5D}\downarrow$ \\ \hline
  &  &  & 18.36 & 446.89 & 22.09 & 450.62 \\
\checkmark &  &  & 16.42 & 441.46 & 21.75 & 446.78 \\
\checkmark & \checkmark &  & 15.78 & 452.05 & 18.59 & 454.86 \\
\checkmark & \checkmark & \checkmark & \textbf{13.92} & \textbf{417.49} & \textbf{16.67} & \textbf{420.24} \\
\Xhline{4\arrayrulewidth}
\end{tabular}
}
\end{adjustbox}
\end{center}
\end{table}

%% file: table/ablation_global_local.tex
\renewcommand{\arraystretch}{1.0}

\begin{table}[t]
\caption{Ablation study on Global and local loss}
\label{table:ablation_global_local}
\begin{center}
\begin{adjustbox}{width=0.45\textwidth}
{
\fontsize{7pt}{9pt}\selectfont
\begin{tabular}{cc|cccc}
\Xhline{3\arrayrulewidth}
\multirow{2}{*}{$L_{global}$} & \multirow{2}{*}{$L_{local}$} & \multicolumn{4}{c}{Performance} \\ \cline{3-6}
&  & $\text{RCD}_{2D}\downarrow$ & $\text{RHD}_{2D}\downarrow$ & $\text{RCD}_{5D}\downarrow$ & $\text{RHD}_{5D}\downarrow$ \\ \hline
\checkmark &  & 15.61 & 441.58 & 18.39 & 444.36 \\
& \checkmark &  \textbf{13.67} & 421.23 & \textbf{16.43} & 424.00 \\
\checkmark & \checkmark & 13.92 & \textbf{417.49} & 16.67 & \textbf{420.24} \\
\Xhline{3\arrayrulewidth}
\end{tabular}
}
\end{adjustbox}
\end{center}
\end{table}

%% file: table/ablation_detection.tex
\renewcommand{\arraystretch}{1.0}

\begin{table}[t]
\caption{detection performance comparison on our self-collected dataset. D: BEV object detection, G: Point generation, Veh: Vehicle, L-Veh: Large Vehicle, Ped: Pedestrian.}
\label{table:ablation_detection}
\begin{center}
\begin{adjustbox}{width=0.45\textwidth}
{
\fontsize{5pt}{7pt}\selectfont
\begin{tabular}{c c 'c c c'c}
\Xhline{2.5\arrayrulewidth} 
Task & Input & \( \text{Veh.} \uparrow \) & \( \text{L-Veh.} \uparrow \) & \( \text{Ped.} \uparrow \) & \( \text{mAP} \uparrow \) \\[2pt]
\hline 
\Xhline{0.1\arrayrulewidth}

D & Low-res & 74.54 & 31.22 & 18.08 & 41.28 \\
D & High-res  & 79.32 & 37.13 & 25.33 & 47.26 \\ 
G+D & Low-res  & 76.00 & 36.82 & 20.49 & 44.44 \\
\Xhline{0.1\arrayrulewidth}
\multicolumn{2}{c'}{Performacne gain} & \textbf{+1.46} & \textbf{+5.60} & \textbf{+2.41} & \textbf{+3.16} \\

\Xhline{2.5\arrayrulewidth}
\end{tabular}

}
\end{adjustbox}
\end{center}
\end{table}

%% file: main.bbl
\begin{thebibliography}{10}

\bibitem{achlioptas2018learning}
Panos Achlioptas, Olga Diamanti, Ioannis Mitliagkas, and Leonidas Guibas.
\newblock Learning representations and generative models for 3d point clouds.
\newblock In {\em International conference on machine learning}, pages 40--49. PMLR, 2018.

\bibitem{neural_LT}
Dzmitry Bahdanau, Kyunghyun Cho, and Yoshua Bengio.
\newblock Neural machine translation by jointly learning to align and translate.
\newblock {\em arXiv preprint arXiv:1409.0473}, 2014.

\bibitem{nuscenes}
Holger Caesar, Varun Bankiti, Alex~H Lang, Sourabh Vora, Venice~Erin Liong, Qiang Xu, Anush Krishnan, Yu~Pan, Giancarlo Baldan, and Oscar Beijbom.
\newblock nuscenes: A multimodal dataset for autonomous driving.
\newblock In {\em Proceedings of the IEEE/CVF Conference on Computer Vision and Pattern Recognition}, pages 11621--11631, 2020.

\bibitem{chen2023anchorformer}
Zhikai Chen, Fuchen Long, Zhaofan Qiu, Ting Yao, Wengang Zhou, Jiebo Luo, and Tao Mei.
\newblock Anchorformer: Point cloud completion from discriminative nodes.
\newblock In {\em Proceedings of the IEEE/CVF Conference on Computer Vision and Pattern Recognition}, pages 13581--13590, 2023.

\bibitem{stargan_I2I}
Yunjey Choi, Minje Choi, Munyoung Kim, Jung-Woo Ha, Sunghun Kim, and Jaegul Choo.
\newblock Stargan: Unified generative adversarial networks for multi-domain image-to-image translation.
\newblock In {\em Proceedings of the IEEE conference on computer vision and pattern recognition}, pages 8789--8797, 2018.

\bibitem{dai2017shape}
Angela Dai, Charles Ruizhongtai~Qi, and Matthias Nie{\ss}ner.
\newblock Shape completion using 3d-encoder-predictor cnns and shape synthesis.
\newblock In {\em Proceedings of the IEEE conference on computer vision and pattern recognition}, pages 5868--5877, 2017.

\bibitem{fan2017point}
Haoqiang Fan, Hao Su, and Leonidas~J Guibas.
\newblock A point set generation network for 3d object reconstruction from a single image.
\newblock In {\em Proceedings of the IEEE conference on computer vision and pattern recognition}, pages 605--613, 2017.

\bibitem{gadelha2018multiresolution}
Matheus Gadelha, Rui Wang, and Subhransu Maji.
\newblock Multiresolution tree networks for 3d point cloud processing.
\newblock In {\em Proceedings of the European Conference on Computer Vision (ECCV)}, pages 103--118, 2018.

\bibitem{han2017high}
Xiaoguang Han, Zhen Li, Haibin Huang, Evangelos Kalogerakis, and Yizhou Yu.
\newblock High-resolution shape completion using deep neural networks for global structure and local geometry inference.
\newblock In {\em Proceedings of the IEEE international conference on computer vision}, pages 85--93, 2017.

\bibitem{han2019multi}
Zhizhong Han, Xiyang Wang, Yu-Shen Liu, and Matthias Zwicker.
\newblock Multi-angle point cloud-vae: Unsupervised feature learning for 3d point clouds from multiple angles by joint self-reconstruction and half-to-half prediction.
\newblock In {\em 2019 IEEE/CVF International Conference on Computer Vision (ICCV)}, pages 10441--10450. IEEE, 2019.

\bibitem{Cond_I2I}
Phillip Isola, Jun-Yan Zhu, Tinghui Zhou, and Alexei~A Efros.
\newblock Image-to-image translation with conditional adversarial networks.
\newblock In {\em Proceedings of the IEEE conference on computer vision and pattern recognition}, pages 1125--1134, 2017.

\bibitem{GrifNet}
Youngseok Kim, Jun~Won Choi, and Dongsuk Kum.
\newblock Grif net: Gated region of interest fusion network for robust 3d object detection from radar point cloud and monocular image.
\newblock In {\em 2020 IEEE/RSJ International Conference on Intelligent Robots and Systems (IROS)}, pages 10857--10864. IEEE, 2020.

\bibitem{adam}
Diederik~P Kingma and Jimmy Ba.
\newblock Adam: A method for stochastic optimization.
\newblock {\em arXiv preprint arXiv:1412.6980}, 2014.

\bibitem{pointpillars}
Alex~H Lang, Sourabh Vora, Holger Caesar, Lubing Zhou, Jiong Yang, and Oscar Beijbom.
\newblock Pointpillars: Fast encoders for object detection from point clouds.
\newblock In {\em Proceedings of the IEEE/CVF conference on computer vision and pattern recognition}, pages 12697--12705, 2019.

\bibitem{pu-gan}
Ruihui Li, Xianzhi Li, Chi-Wing Fu, Daniel Cohen-Or, and Pheng-Ann Heng.
\newblock Pu-gan: a point cloud upsampling adversarial network.
\newblock In {\em Proceedings of the IEEE/CVF international conference on computer vision}, pages 7203--7212, 2019.

\bibitem{dis-pu}
Ruihui Li, Xianzhi Li, Pheng-Ann Heng, and Chi-Wing Fu.
\newblock Point cloud upsampling via disentangled refinement.
\newblock In {\em Proceedings of the IEEE/CVF conference on computer vision and pattern recognition}, pages 344--353, 2021.

\bibitem{pointcnn}
Yangyan Li, Rui Bu, Mingchao Sun, Wei Wu, Xinhan Di, and Baoquan Chen.
\newblock Pointcnn: Convolution on x-transformed points.
\newblock {\em Advances in neural information processing systems}, 31, 2018.

\bibitem{focal}
Tsung-Yi Lin, Priya Goyal, Ross Girshick, Kaiming He, and Piotr Doll{\'a}r.
\newblock Focal loss for dense object detection.
\newblock In {\em Proceedings of the IEEE international conference on computer vision}, pages 2980--2988, 2017.

\bibitem{pc2-pu}
Chen Long, WenXiao Zhang, Ruihui Li, Hao Wang, Zhen Dong, and Bisheng Yang.
\newblock Pc2-pu: Patch correlation and point correlation for effective point cloud upsampling.
\newblock In {\em Proceedings of the 30th ACM International Conference on Multimedia}, pages 2191--2201, 2022.

\bibitem{luo2021diffusion}
Shitong Luo and Wei Hu.
\newblock Diffusion probabilistic models for 3d point cloud generation.
\newblock In {\em Proceedings of the IEEE/CVF Conference on Computer Vision and Pattern Recognition}, pages 2837--2845, 2021.

\bibitem{effective_LT}
Minh-Thang Luong, Hieu Pham, and Christopher~D Manning.
\newblock Effective approaches to attention-based neural machine translation.
\newblock {\em arXiv preprint arXiv:1508.04025}, 2015.

\bibitem{nguyen2016field}
Duc~Thanh Nguyen, Binh-Son Hua, Khoi Tran, Quang-Hieu Pham, and Sai-Kit Yeung.
\newblock A field model for repairing 3d shapes.
\newblock In {\em Proceedings of the IEEE Conference on Computer Vision and Pattern Recognition}, pages 5676--5684, 2016.

\bibitem{nie2020skeleton}
Yinyu Nie, Yiqun Lin, Xiaoguang Han, Shihui Guo, Jian Chang, Shuguang Cui, Jian Zhang, et~al.
\newblock Skeleton-bridged point completion: From global inference to local adjustment.
\newblock {\em Advances in Neural Information Processing Systems}, 33:16119--16130, 2020.

\bibitem{pointnet}
Charles~R Qi, Hao Su, Kaichun Mo, and Leonidas~J Guibas.
\newblock Pointnet: Deep learning on point sets for 3d classification and segmentation.
\newblock In {\em Proceedings of the IEEE conference on computer vision and pattern recognition}, pages 652--660, 2017.

\bibitem{pointnet++}
Charles~Ruizhongtai Qi, Li~Yi, Hao Su, and Leonidas~J Guibas.
\newblock Pointnet++: Deep hierarchical feature learning on point sets in a metric space.
\newblock {\em Advances in neural information processing systems}, 30, 2017.

\bibitem{pu-gcn}
Guocheng Qian, Abdulellah Abualshour, Guohao Li, Ali Thabet, and Bernard Ghanem.
\newblock Pu-gcn: Point cloud upsampling using graph convolutional networks.
\newblock In {\em Proceedings of the IEEE/CVF Conference on Computer Vision and Pattern Recognition}, pages 11683--11692, 2021.

\bibitem{pugeo-net}
Yue Qian, Junhui Hou, Sam Kwong, and Ying He.
\newblock Pugeo-net: A geometry-centric network for 3d point cloud upsampling.
\newblock In {\em European conference on computer vision}, pages 752--769. Springer, 2020.

\bibitem{pu-transformer}
Shi Qiu, Saeed Anwar, and Nick Barnes.
\newblock Pu-transformer: Point cloud upsampling transformer.
\newblock In {\em Proceedings of the Asian Conference on Computer Vision}, pages 2475--2493, 2022.

\bibitem{SDF_diffusion}
Jaehyeok Shim, Changwoo Kang, and Kyungdon Joo.
\newblock Diffusion-based signed distance fields for 3d shape generation.
\newblock In {\em Proceedings of the IEEE/CVF Conference on Computer Vision and Pattern Recognition}, pages 20887--20897, 2023.

\bibitem{shu20193d}
Dong~Wook Shu, Sung~Woo Park, and Junseok Kwon.
\newblock 3d point cloud generative adversarial network based on tree structured graph convolutions.
\newblock In {\em Proceedings of the IEEE/CVF international conference on computer vision}, pages 3859--3868, 2019.

\bibitem{sun2020pointgrow}
Yongbin Sun, Yue Wang, Ziwei Liu, Joshua Siegel, and Sanjay Sarma.
\newblock Pointgrow: Autoregressively learned point cloud generation with self-attention.
\newblock In {\em Proceedings of the IEEE/CVF Winter Conference on Applications of Computer Vision}, pages 61--70, 2020.

\bibitem{S2S_LT}
Ilya Sutskever, Oriol Vinyals, and Quoc~V Le.
\newblock Sequence to sequence learning with neural networks.
\newblock {\em Advances in neural information processing systems}, 27, 2014.

\bibitem{tang2022lake}
Junshu Tang, Zhijun Gong, Ran Yi, Yuan Xie, and Lizhuang Ma.
\newblock Lake-net: Topology-aware point cloud completion by localizing aligned keypoints.
\newblock In {\em Proceedings of the IEEE/CVF conference on computer vision and pattern recognition}, pages 1726--1735, 2022.

\bibitem{valsesia2018learning}
Diego Valsesia, Giulia Fracastoro, and Enrico Magli.
\newblock Learning localized generative models for 3d point clouds via graph convolution.
\newblock In {\em International conference on learning representations}, 2018.

\bibitem{vaswani2017attention}
Ashish Vaswani, Noam Shazeer, Niki Parmar, Jakob Uszkoreit, Llion Jones, Aidan~N Gomez, {\L}ukasz Kaiser, and Illia Polosukhin.
\newblock Attention is all you need.
\newblock {\em Advances in neural information processing systems}, 30, 2017.

\bibitem{wang2021voxel}
Xiaogang Wang, Marcelo~H Ang, and Gim~Hee Lee.
\newblock Voxel-based network for shape completion by leveraging edge generation.
\newblock In {\em Proceedings of the IEEE/CVF international conference on computer vision}, pages 13189--13198, 2021.

\bibitem{dgcnn}
Yue Wang, Yongbin Sun, Ziwei Liu, Sanjay~E Sarma, Michael~M Bronstein, and Justin~M Solomon.
\newblock Dynamic graph cnn for learning on point clouds.
\newblock {\em ACM Transactions on Graphics (tog)}, 38(5):1--12, 2019.

\bibitem{wen2021cycle4completion}
Xin Wen, Zhizhong Han, Yan-Pei Cao, Pengfei Wan, Wen Zheng, and Yu-Shen Liu.
\newblock Cycle4completion: Unpaired point cloud completion using cycle transformation with missing region coding.
\newblock In {\em Proceedings of the IEEE/CVF conference on computer vision and pattern recognition}, pages 13080--13089, 2021.

\bibitem{wen2020point}
Xin Wen, Tianyang Li, Zhizhong Han, and Yu-Shen Liu.
\newblock Point cloud completion by skip-attention network with hierarchical folding.
\newblock In {\em Proceedings of the IEEE/CVF conference on computer vision and pattern recognition}, pages 1939--1948, 2020.

\bibitem{xie2020grnet}
Haozhe Xie, Hongxun Yao, Shangchen Zhou, Jiageng Mao, Shengping Zhang, and Wenxiu Sun.
\newblock Grnet: Gridding residual network for dense point cloud completion.
\newblock In {\em European Conference on Computer Vision}, pages 365--381. Springer, 2020.

\bibitem{UltraLiDAR}
Yuwen Xiong, Wei-Chiu Ma, Jingkang Wang, and Raquel Urtasun.
\newblock Learning compact representations for lidar completion and generation.
\newblock In {\em Proceedings of the IEEE/CVF Conference on Computer Vision and Pattern Recognition}, pages 1074--1083, 2023.

\bibitem{yang2019pointflow}
Guandao Yang, Xun Huang, Zekun Hao, Ming-Yu Liu, Serge Belongie, and Bharath Hariharan.
\newblock Pointflow: 3d point cloud generation with continuous normalizing flows.
\newblock In {\em Proceedings of the IEEE/CVF international conference on computer vision}, pages 4541--4550, 2019.

\bibitem{yang2018foldingnet}
Yaoqing Yang, Chen Feng, Yiru Shen, and Dong Tian.
\newblock Foldingnet: Point cloud auto-encoder via deep grid deformation.
\newblock In {\em Proceedings of the IEEE conference on computer vision and pattern recognition}, pages 206--215, 2018.

\bibitem{mpu}
Wang Yifan, Shihao Wu, Hui Huang, Daniel Cohen-Or, and Olga Sorkine-Hornung.
\newblock Patch-based progressive 3d point set upsampling.
\newblock In {\em Proceedings of the IEEE/CVF Conference on Computer Vision and Pattern Recognition}, pages 5958--5967, 2019.

\bibitem{pu-net}
Lequan Yu, Xianzhi Li, Chi-Wing Fu, Daniel Cohen-Or, and Pheng-Ann Heng.
\newblock Pu-net: Point cloud upsampling network.
\newblock In {\em Proceedings of the IEEE conference on computer vision and pattern recognition}, pages 2790--2799, 2018.

\bibitem{yu2021pointr}
Xumin Yu, Yongming Rao, Ziyi Wang, Zuyan Liu, Jiwen Lu, and Jie Zhou.
\newblock Pointr: Diverse point cloud completion with geometry-aware transformers.
\newblock In {\em Proceedings of the IEEE/CVF international conference on computer vision}, pages 12498--12507, 2021.

\bibitem{yuan2018pcn}
Wentao Yuan, Tejas Khot, David Held, Christoph Mertz, and Martial Hebert.
\newblock Pcn: Point completion network.
\newblock In {\em 2018 international conference on 3D vision (3DV)}, pages 728--737. IEEE, 2018.

\bibitem{Nerf-lidar}
Junge Zhang, Feihu Zhang, Shaochen Kuang, and Li~Zhang.
\newblock Nerf-lidar: Generating realistic lidar point clouds with neural radiance fields.
\newblock {\em arXiv preprint arXiv:2304.14811}, 2023.

\bibitem{zhou2022seedformer}
Haoran Zhou, Yun Cao, Wenqing Chu, Junwei Zhu, Tong Lu, Ying Tai, and Chengjie Wang.
\newblock Seedformer: Patch seeds based point cloud completion with upsample transformer.
\newblock In {\em European conference on computer vision}, pages 416--432. Springer, 2022.

\bibitem{Unpaired_I2I}
Jun-Yan Zhu, Taesung Park, Phillip Isola, and Alexei~A Efros.
\newblock Unpaired image-to-image translation using cycle-consistent adversarial networks.
\newblock In {\em Proceedings of the IEEE international conference on computer vision}, pages 2223--2232, 2017.

\bibitem{zyrianov2022learning}
Vlas Zyrianov, Xiyue Zhu, and Shenlong Wang.
\newblock Learning to generate realistic lidar point clouds.
\newblock In {\em European Conference on Computer Vision}, pages 17--35. Springer, 2022.

\end{thebibliography}
